\pdfoutput=1
\documentclass[10pt,journal,compsoc]{IEEEtran}

\ifCLASSOPTIONcompsoc
  \usepackage[nocompress]{cite}
\else
  \usepackage{cite}
\fi
\ifCLASSINFOpdf
\else
\fi
\usepackage{amsmath}
\usepackage{amssymb}
\newtheorem{theorem}{Theorem}
\newtheorem{corollary}{Corollary}[theorem]

\newtheorem{definition}{Definition}
\usepackage{algorithm}
\usepackage{algpseudocode}

\begin{document}
%
\title{Active Robust Learning}
%
%
%
%

\author{Hossein~Ghafarian,
        and~Hadi~Sadoghi Yazdi
\IEEEcompsocitemizethanks{\IEEEcompsocthanksitem M. Hossein Ghafarian was with the Department
of Computer Engineering, Ferdowsi University of Mashhad, Iran. \protect\\
E-mail: see s.h.ghafarian@gmail.com
\IEEEcompsocthanksitem with the Department
of Computer Engineering, Ferdowsi University of Mashhad, Iran.}
\thanks{Manuscript August 1, 2016}}

%
%

\markboth{}%
{Shell \MakeLowercase{\textit{et al.}}: Bare Demo of IEEEtran.cls for Computer Society Journals}
%



\IEEEtitleabstractindextext{%
\begin{abstract}
In many practical applications of learning algorithms, unlabeled data is cheap and abundant whereas labeled data is expensive. Active learning algorithms developed to achieve better performance with lower cost. Usually Representativeness and Informativeness are used in active learning algoirthms. Advanced recent active learning methods consider both of these criteria. Despite its vast literature, very few active learning methods consider noisy instances, i.e. label noisy and outlier instances. Also, these methods didn't consider accuracy in computing representativeness and informativeness. Based on the idea that inaccuracy in these measures and not taking noisy instances into consideration are two sides of a coin and are inherently related, a new loss function is proposed. This new loss function helps to decrease the effect of noisy instances while at the same time, reduces bias. We defined "instance complexity" as a new notion of complexity for instances of a learning problem. It is proved that noisy instances in the data if any, are the ones with maximum instance complexity. Based on this loss function which has two functions for classifying ordinary and noisy instances, a new classifier, named "Simple-Complex Classifier" is proposed. In this classifier there are a simple and a complex function, with the complex function responsible for selecting noisy instances. The resulting optimization problem for both learning and active learning is highly non-convex and very challenging. In order to solve it, a convex relaxation is proposed. In every iteration of active learning, a problem with some small changes needs to be solved. In order to take the advantage of this, an algorithm for solving this problem is proposed which is capable of using the most information available from the previous solutions of the problem. Accelerated version of the optimization algorithm is also proposed. Theoretical and experimental studies show that this method is efficient.  
\end{abstract}

}

\maketitle

\IEEEdisplaynontitleabstractindextext

%
\IEEEpeerreviewmaketitle

\IEEEraisesectionheading{\section{Introduction}\label{sec:introduction}}

%
%
%
%
\IEEEPARstart{S}{upervised} machine learning  methods need labeled data. Labeled information is expensive and human annotators often provide noisy labels. On the other hand, unlabeled data is cheap and easy to obtain. To reduce the cost of building efficient learning systems, Semi-supervised and Active Learning methods have been developed. In active learning, it is assumed that knowledge about labels of some data points are more usefull than others\cite{Hoi2009}. Also, semi-supervised learning methods assume that unlabeled data has knowledge which can be utilised for learning. 

Unfortunately, these methods sometimes encountered degraded performance\cite{Li2014,Hanneke2015}. One of its reasons is noisy instances. Noisy instances can be outlier or label noisy. These data points largely affect classfication boundary and reduce the generalizability of resulting classifier. In spite of this, there are few reasearches about noisy data in active learning\cite{Zhang2015}. 

Recently, Hanneke \cite{Hanneke2015} showed that in active learning with convex surrogate losses achieving maximum improvment in sample complexity, i.e. number of queries necessary for a certain accuracy, in presence of noise is not possible. This is because noisy data points move classifier boundary away from optimial by having large impact on it and this results to more queries. In active learning especially in early stages there are very few labeled instances. This makes active learning methods more vulnerable to noisy instances. Moreover, correctly labeled outlier instances may reduce generalizability of classifier\cite{Zhang2015}. With incomplete information in active learning applications, it's not possible to remove instances considered noisy in each stage. By doing this, we may lose some portions of boundary, since with new labels, these instances may be no longer considered noisy.

In this paper, we argue that in order to have more effective learning and active learning algorithms in presence of noisy instances, we must have a noise resistant mechanism at the classifier level. We propose a method that under mild conditions can remove effect of both outlier and noisy label instances as much as possible. It is proved that this method is unbiased. 

Based on this method, Robust Active Learning (RAL) is devised to minimize the effect of noisy instances on both the classifier boundary and sample complexity. In semi-supervised learning and active learning, where all the data is not labeled, learning can easily get biased. This is more severe in active learning, which labeled data are no longer i.i.d. In RAL unlabeled data is used in order to alleviate this problem.  

Informativeness and representativeness are two criteria which many active learning methods use. RAL considers both of these criteria. In addition, by paying attention to noisy instances and reducing effects of these instances through an unbiased method, RAL also concentrates on the accuracy of these criteria. 

The intution behind the proposed method is that usefull data for learning lies in a range of complexities. We define the notion of instance complexity as the amount of distortion it can be added without changing of learning function much. For noisy instances, even a small distortion makes the learning function much different. Also, for noisy labeled instances a small change  of instance may result in quite a change of learning function. Non-noisy instances on the other hand are more resistant to distortion. The notion of instance complexity is defined based on the equivanlence of robustness and regularization\cite{Xu2009}. 
 
In order to discriminate noisy instances we devise a new loss functions. For noisy instances loss function value is small and for other instances its value is equal to ordinary loss values. In this loss function two functions are involved. A simpler function corresponding to ordinary classification function and a more complex function. This complex function is used to discriminate noisy instances. 

The resulting optimization problem is highly non-convex and very challenging. A convex-relaxation for RAL is developed to obtain a good approximation to solution of the problem. (we show that objective of this approxiamtion is near to global optimal solution of the original non-convex problem).

Main contributions of the paper are as follows: 
\begin{itemize}
	\item Definition of instance complexity based on distortion of noisy instances 
	\item The new loss function which is robust to outlier and noisy labeled instances
	\item Convex Relaxation for Robust Active Learning
	\item Efficient Algorithm for solving Convex Relaxation problem capable of warm-starting effectively which is necessary for Active Learning.  
\end{itemize}
 
In the following, we first review related works. After that notation and some preliminaries is dicussed. In section 2, problem formulation is stated. In section 3, Simple-Complex classifier, problem is dicussed. In the next section, some theoretical results regarding the simple-complex classifier is introduced. In section 5, RAL convex relaxation is solved using a Nesterov's method. In section 6, experimental results are analysed. And finally, future works and conclusion is in the last section.  
\subsection{Motivation and Related Works}
Someone may think why not eliminate outlier instances before active learning began? Before starting and in the first stages of active learning, we have very few labels. Therefore, an accurate enough estiamte of classifier is not accessible. In this situation by elimination of data we may lose some important information. Furthermore, in some applications we may not have an accurate estimate of the instances themselves. For example, in learning from distributions, when every instance itself is a distribution, many modern methods use embedding of distributions on a reproducing kernel hilbert space. Usually, there are a small sample drawn from a distribution and estimating kernel embedding for distribution is not accurate\cite{Muandet2016}. When we don't have an accuate estimate of both learning function and instances, eliminating precious information may not be a good choice. 

Also, in some applications, we need to relearn the function with the newly arrived data. Unless we have plenty of data, we don't know in advance whether some data is outlier or not, yet we have to learn using existing data. 

Depending on the degree of outlyingness\cite{Dunagan2002}, eliminating instances in early stages of learning causes different degrees of risk. If degree of outlyingness for an instance is very high, it may be safe to delete it. But if degree of outlyingness is not very high deleting instance from data, will cause considerable risk to learning.

In many active learning application, aquiring label information is highly expensive. For example, when labels come from an expensive or time consuming experiment or they come from expensive experts time. In some other cases, it may be very important to have highest possible accuracy, using available data,(e.g. Mine fields prediction). In such case, it is very important to use all means for achieving a more accurate classifier using as few as possible labeled data and therefore, may be speed of querying is not so imporant.
  
Active Learning for Support Vecctor Machines is introduced in \cite{Tong2001}. Their approach is to select data point that minimizes version space. For support vector machines this means selecting data point nearest to current boundary. Extending this approach to selecting a batch of data points is difficult. 

Based on Active Learning for Support Vector Machines, min-max framework for Active learning is proposed\cite{Hoi2009a}. In this framework, nearest point to boundary regardless of its label is selected. When a new labeled instance added, assuming classifier is fixed results to over-estimation of the impact of this new labeled point\cite{Hoi2009a}. Therefore, we must simultaneously optimize on both classifier and query point. Notice that the query point is no longer the nearest point to current classification boundary. This approach selects the most \textit{informative} instance. Informativeness of a data point is measured with uncertainty of predicting its label using labeled data.  

Selecting most informative instances makes learning biased to sample. To solve this issue usually \textit{representativeness} is also considered. Many active learning methods are adhoc in combining repesentativeness and informativeness. In \cite{Huanga} representativeness of an example measured by uncertainty in predicting its label based on unlabeled data in min-max framework\cite{Hoi2009a}. Minimizing svm cost function with respect to classifier, unlabeled data and selection variable for active learning, makes the problem very challenging. Instead \cite{Huanga} used least squares  as loss function. Objective function value with least squares loss has a close form, which they exploit to compute query point. Although they used least squares loss function in active learning, they used svm for computing classifier to report accuracy. In classification, hinge loss considered a better surrogate for 0-1 loss than least squares. Although a different loss in active learning and learning used which may bias the model, they didn't prove any results regarding unbiasedness of the learning. In our method we used hinge loss for both classificatin and active learning and solved a semi-supervised learning in every iteration. In this way better measures of informativeness and representativeness is obtained.         

\cite{Zhang2015} states that outlier and label noisy data are harmfull for active learning and proposed a forward-backward approach to explore unlabeled and labeled data to get rid off them. In forward phase they query to add unlabeled data to labeled dataset. In backward phase, some most probable noisy instances are removed from labeled dataset to protect classifier from their impact. These data points deteriorate classifier, the most. Decision on which data points are noisy performed on two levels, instance and label-level. 

The authors in \cite{Chakraborty2015} proposed a convex-relaxation for active learning. They first constructed a matrix based on the uncertainty and divergence of data and then selects some rows/columns of matrix using an integer quadrtic program, which they managed to develop a convex-relaxation for. As stated \cite{Huanga}, this kind of combining informativness and representativeness are adhoc and a more principled approach is usually prefered. 

Regarding instace complexity, same as \cite{Xu2009}, it is different from Influence function(IF). IF considers the change of functions from the perspective of classifier\cite{Christmann2004}. On the other hand, instance complexity considers change of classifier by disturbances in an instance from the perspective of instance. 

In \cite{Dunagan2002} a geometric theory for outliers developed, which we used their definition.

\section{Active and Semi-supervised Learning}
Min-max framework to active learning \cite{Hoi2009a,Huanga} attempts to find the most usefull instances using labeled and unlabeled data. Unfortunately, noisy labeled data can impact classification boundary severly. Also, it's well known that classifier may overfit to correctly labeled outliers\cite{Zhang2015}. Therefore, noisy data may decrease generalization. In order to improve accuracy in presence of noisy training data, the following scheme is proposed. 
\subsection{The Framework}
Let $\mathcal{S}_0= \{(x_i,y_i)|x_i\in \mathbb{R}^d,y_i \in \{-1,1\}\}$ be the non-noisy initial labeled set. Define cost function as 
\begin{IEEEeqnarray}{rcl}
J(f,\lambda)=\sum_{i\in \mathcal{S}_0} l(y_{i},f(x_i))+\lambda 
\Vert f\Vert ^{2}
\IEEEeqnarraynumspace
\label{eq:labelc1}
\end{IEEEeqnarray} 
Let $f^*$ be the optimal classifier. Version space minimization approach select closest instance to classification boundary, i.e, $x_s^* = arg\min_{x_s} \vert f^*(x_s)\vert$. Unfortnately, especially in early stages of active learning, current boundary differ much from optimal boundary. By increasing number of labeled instances, it is expected that current boundary moves closer to optimal boundary. Based on \cite{Hoi2009a,Huanga}, cost function of minimax framework for active learning, can be written as
\begin{align}
\begin{split}
x^*_s&=argmin_{x_s}\min _{\lbrace J(f,\lambda ^{*})<= 
	J(f^{*},\lambda )\rbrace }\vert f(x_s)\vert \\
&=argmin_{x_s }\min _{f\varepsilon 
	H}c_{s}|f(x_s)|+J(f,\lambda ^{}) \\
&=argmin_{x_s }\max _{y_{s}}\min _{f\varepsilon H} 
c_{s}l(y_{s},f(x_s))+J(f,\lambda ^{})\\
\end{split}
\end{align}
In the second equality, $c_s$ is depenedent to $f(x_s)$ and $J(f,\lambda)$. Since there may be noisy instances, we cannot assume that this coefficients are equal for every point. Using $c_s$, importance of loss function for instance $x_s$ could be adjusted compared to empirical risk of other instances. A small $c_s$ means that this data point is not important for learning the function. For representativeness\cite{Huanga} of querypoints, minimization on lables of unlabeled data is used, i.e.
\begin{IEEEeqnarray}{l}
	\label{alfirstobjective}
	x_s=arg\min_{x_s} \min_{y_u} \max_{y_s}   \mathcal{L}(\mathcal{D-S}_0,x_s) \label{eq:alproblem} \\ 
	\mathcal{L}(\mathcal{U},x_s)= \min_{f\in H}\sum _{i\in \mathcal{U}}  c_i l(y_{i},f(x^{i}))+ J(f,\lambda)+\frac{\lambda}{2}\Vert f\Vert ^{2} 
	\label{eq:alcostfunction}
\end{IEEEeqnarray}  
Unfortunately, the coefficients $c_i$ for all unlabeled instaces are unknown. But we know that it is zero for noisy instances. Noisy instance's impact on the classifier is undesireable. Let $\mathcal{D}_o$ be the set of noisy instaces, we can simply set
\begin{align}\label{eq:ciloss}
c_i=\begin{cases}
1   &i\in \mathcal{D}-\mathcal{D}_o\\
0   &i\in \mathcal{D}_o
\end{cases}\notag
\end{align}
Assume we have access to a function $f_o(x_i)$ with value zero for non-noisy instances and $y_i$, i.e. hypothetical class label of noisy instances. 
\begin{align*}
f_o(x_i)=
\begin{cases}
0     &i\in \mathcal{D}-\mathcal{D}_o\\
y_i   &i\in \mathcal{D}_o
\end{cases}\notag
\end{align*}
If including loss for an instance $x_i$ in cost function is usefull, $f_o(x_i)$ must be zero. Also if function $f_o$ assigns a value other than zero to an instance, loss of this instance must be zero. we can set $c_i=l_{0,1}(y_i f_o(x_i))$ using the loss
\begin{align*}
l_{0,1}(u)=\begin{cases}
1   &u = 0\\
0   &u \in\{-1,1\} 
\end{cases}\notag
\end{align*}  

\begin{corollary}
	With $c_i$ defined as above, problem (\ref{eq:alproblem}) is equaivalent with the same problem when there are no noisy instances.
\end{corollary}
Unfortunately noisy data set $\mathcal{D}_o$ as well as function $f_o$ are unknown. How it's possible to remove those instances from learning when they are unkown? 
\section{Simple-Complex Classifier}
In order to provide an answer to this question, we first define notion of instance complexity. This defintion is motivated by equivalence of regualrization and robustness\cite{Xu2009}. Let $\mathcal{D}_i^\delta = \mathcal{D}-{(x_i,y_i)}\cup {(x_i+\delta,y_i)}$ , $f_i^0 = f$ and $f_i^\delta=arg\min_f \sum_{j\in \mathcal{D}_i^\delta} l(y_j f(x_j))+
\frac{\lambda}{2}\Vert f\Vert^2$.
\begin{definition}{Instance Complexity}
	Define instance complexity of instance $x_i$, $c(x_i)$ as 
	 \begin{IEEEeqnarray}{rcl}
	 	c(x_i)=&\min_{d}\quad &d^{-1}\\
		&s.t. &\forall \delta, \Vert \delta\Vert \leq d, \Vert f_i^\delta-f\Vert \leq \epsilon\IEEEnonumber
	 \end{IEEEeqnarray}
\end{definition}
This definition is very intuitive. If instance $x_i$ is simple, changing it to $x_i+\delta$ with $\Vert \delta\Vert$ large doesn't change function learned on this data much. But even a small disturbation on a complex instance makes learnt function much different. In otherwords, classifier is too sensitive to complex instances. 
The following therorem proves that if there are any noisy instances, they are the most compelx instances. 
\begin{theorem}
	\label{instanceComplexityTheorem}
	Assume $f$ learned on $\mathcal{D}$ which has a subset of the size $n_o$ of noisy instances. Then these instances have the highest $n_o$ instance complexity values $c(x_i)$.  
\end{theorem}

Based on theorem  (\ref{instanceComplexityTheorem}), we need a mechanism to select the most complex instances. As stated before, usefull instances for learning are in a range of complexities. Too simple instaces are not usefull for learning and too complex instances are noisy and harmfull for learning. Therefore, more complex instances must be classified using function $f_o$ and simple instances using function $f$. In order to restrict classifier $f_o$ from classifying simple instances,  total energy of this function must be limited. In otherwords, since there is only a limited amount of noisy data, number of non-zero values of $f_o$ must be limited. Therefore with $f_o$ being complex enough and more complex than $f$, noisy instances will be selected by $f_o$. Considering the constraint $f_o(x_i)=0$ for initially labeled instances and assuming there is only $n_o$ noisy instances, problem (\ref{eq:alcostfunction}) just becomes: 
\begin{IEEEeqnarray}{lcl}
	\mathcal{L}(\mathcal{D},x_s)= &\min_{f\in \mathcal{H},f_o\in\mathcal{H}_o} &\sum _{i\in \mathcal{D}}  l_{0,1}(y_if_o(x_i)) l(y_{i}f(x^{i}))\IEEEnonumber\\
	&&+\frac{\lambda}{2}\Vert f\Vert_{\mathcal{H}} ^{2}+\frac{\lambda_o}{2}\Vert \label{l01problem} f_o\Vert_{\mathcal{H}_o} ^{2}\IEEEnonumber\\
	&s.t.&  \sum   \vert f_o(x_i) \vert \leq n_o\IEEEnonumber\\
	&&\vert f_o(x_i) \vert \leq 1 
\end{IEEEeqnarray} 
We can adjust complexity of function $f_o$ with properties of reproducing kernel hilbert space, $\mathcal{H}_o$ and parameter $\lambda_o$. 

Replacing $c_i$ with $l_{0,1}(y_if_o(x_i))$ is very intuitive. In Theoretical Results section, we prove that this cost function finds noisy instances. Furthermore, it is proven that this cost functions gives us an unbiased classifier.
Let $l_i=l(y_if(x_i))$ and $l_i^o=l_{0,1}(y_if_o(x_i))$. It is very inuitive that in ideal case, we want loss vectors $l=(l_1,l_2,....l_n)$ and $l_o=(l_1^o,l_2^o,....l_n^o)$ to be orthogonal.  

Loss function $l_{0,1}(.)$ is non-convex. If we assume $y_if_o(x_i)\ge 0$, then $l_{0,1}$ can be approximated by any surrogate loss function such as hinge loss. Also, in the following $l$ is used as hinge. Let $f(x_i)=\langle w,\phi(x_i)\rangle$ and $f_o(x_i)=\langle w_o,\psi(x_i)\rangle$, where $\phi(x_i)\in \mathcal{H}$ and $\psi(x_i)\in \mathcal{H}_o$. Replacing both losses with hinge loss, we reach to the following problem:
\begin{IEEEeqnarray}{lcl}
	\label{Simple-ComplexNonConvexProblem}
	\mathcal{L}(\mathcal{D},x_s)= &\min_{w\in \mathcal{H},w_o\in\mathcal{H}_o} &\sum _{i\in \mathcal{D}}  l_i^o l_i+\frac{\lambda}{2}\Vert w\Vert_{\mathcal{H}} ^{2}+\frac{\lambda_o}{2}\Vert \label{doublehingproblem} w_o\Vert_{\mathcal{H}_o} ^{2}\\
	&s.t.&  \sum   \vert \langle w_o,\psi(x_i)\rangle \vert \leq n_o\\
	&&\vert \langle w_o,\psi(x_i)\rangle \vert \leq 1 
\end{IEEEeqnarray} 
Proof of the following is in Supplementary  Material. 
\begin{theorem}
	\label{superviseSimpleComplexTheorem}
	Let $Y=diag(y)$ and $h=1-Y\Psi(X)^\mathsf{T} w_o$. $\mathcal{L}(\mathcal{D},x_s)$ in problem above is equivalent to 
	\begin{IEEEeqnarray}{lll}
		\min_{w_o,p,G}\max_{\alpha} &\sum_{i\in D} \alpha_i (1-y_{i}w_o^\mathsf{T}\psi(x_i))_+ \IEEEnonumber\\&- \frac{1}{2\lambda} \alpha^\mathsf{T} Y K \odot G Y \alpha +\frac{\lambda_o}{2}\Vert w_o\Vert_{\mathcal{H}_o} ^{2}\\
		 \text{s.t.} &0\leq \alpha \leq 1\\
		& \vert \Psi(X)^\mathsf{T} w_o  \vert\leq p, p \leq 1, p^\mathsf{T} 1 \leq n_o\\
		&G\succeq h h^\mathsf{T}, diag(G)=h\\
		& rank(G)=1
	\end{IEEEeqnarray}
\end{theorem}
The rank constraint makes the above problem nonconvex. Removing it and deriving dual of the inner problem with respect to $\alpha$ similar to \cite{Bie}, the following is obtained. See proof in Supplementary Material.
\begin{theorem}
	\label{superviseSCConvexRelax}
	Convex relaxation for the problem above is 
	\begin{IEEEeqnarray}{lll}
		\label{ConvexRelaxSCProb}
		\min_{w_o,p,G}\min_{\beta^{\prime},\eta^{\prime}} &t+\frac{\lambda_o}{2}\Vert w_o\Vert_{\mathcal{H}_o} ^{2}+\beta^{\prime \mathsf{T}} 1\\
		\text{s.t.}
		&\begin{bmatrix}
			K\odot H &h+ \eta^{\prime}-\beta^{\prime}\\
			(h+ \eta^{\prime}-\beta^{\prime})^\mathsf{T} &\frac{2}{\lambda}t 
		\end{bmatrix} \succeq 0 \\ 
		&\beta^{\prime},\eta^{\prime} \geq 0\\
		& \vert \Psi(X)^\mathsf{T} w_o  \vert \leq p, p \leq 1, p^\mathsf{T} 1 \leq n_o\\
		&\begin{bmatrix}
			H &Yh\\
			(Yh)^\mathsf{T} &1 
		\end{bmatrix} \succeq 0\\
		&diag(H)=1-\Psi(X)^\mathsf{T} w_o
	\end{IEEEeqnarray}
\end{theorem}

\section{Theoretical Results}

In the following theorem based on the definition of outlyingness in \cite{Dunagan2002}, discrimination of noisy instances is considered.
Proof ot this theorem based on a set of lemmas are in Supplementary Material.
\begin{theorem}
	\label{wodiscriminationtheorem}
	If classifier $w$ is simple enough and there exists a direction $w_\beta$ such that it's sign for instance $x_i$, $z_i=sign(w_{\beta}^\mathsf{T}\psi(x_i))$ have enough compliance with instance labes, such that the following inequalities satisfy
	\begin{IEEEeqnarray}{lll}
		\mathbf{1}^\mathsf{T} Z Y l^{*}  \geq \IEEEnonumber\\ 
		\frac{\lambda_o}{m} (\frac{\mu}{2} +  \frac{n_o}{(1+\frac{n-1}{\gamma})/n_s}) \sqrt{1+\frac{n_o}{n_s}} \sqrt{1+\frac{\sum_{t\in D_o}{\gamma_t^2}}{n_s}},\IEEEnonumber\\
		e_{D_o}^\mathsf{T} (\Gamma-I)  Z Y (I -\frac{\mu \lambda_o}{2} B) l^{*} \geq 0 \IEEEnonumber
	\end{IEEEeqnarray}
	where $Z=diag(z)$. Then classifier $w_o$ will discriminate noisy instances.
\end{theorem}
When there is a coefficient in loss function, it's possible that learning becomes bias. In the following, it is proved that under mild conditions, Simple-Complex classifier isn't biased.
\begin{theorem}
	\label{unbiasedtheorem}
	Let $P$ be the distribution data without noise. Let $Q$ be a distribution the same as $P$ but contaminated with noise. Now assume the following condition is satisfied at optimality in (\ref{Simple-ComplexNonConvexProblem}),
	\begin{IEEEeqnarray}{lll}
	\forall i\in \mathcal{S}, &\vert \phi(x_i)^\mathsf{T} w_o^* \vert \leq \epsilon_1\IEEEnonumber\\
	\forall i\in \mathcal{D}_o, &\vert \phi(x_i)^\mathsf{T} w_o^* \vert \geq 1-\epsilon_2
	\end{IEEEeqnarray}
	If $P(y\vert x)$ is fixed, then
	\begin{IEEEeqnarray}{lll}
	\sup_{l(.,.,\theta) \in \mathcal{G}} \Big\vert \frac{1}{n} \sum_{i=1}^n l_{w_o}^* l(x_i,y_i,\theta)- 
	\mathbf{E}_{Y^\prime\vert X^\prime}\Big[ \frac{1}{n^\prime} \sum_{i=1}^{n^\prime} l(x_i^\prime,y_i^\prime,\theta)\Big] \Big\vert  \IEEEnonumber\\ \leq \frac{(1+\sqrt{(-log\delta)/2})2CR}{\sqrt{M}}+C\epsilon_0 \IEEEnonumber
	\end{IEEEeqnarray}
	where $l_{w_o}^*=(1-y_i \phi(x_i)^\mathsf{T} w_o^*)$ , $\epsilon_0=\epsilon  + \frac{1}{n} \sum_{i=1}^{n} ( \epsilon_1 B+\frac{n_o}{n}  (1-\epsilon_1)B)$
\end{theorem}
This result proved based on \cite{Huang2007} in Supplementary  Material ,
shows that if weighted risk function based on Simple-Complex classifier is minimized, with high probability, an upper bound of expected risk on test sample is minimized. This is very interesting. This shows that Simple-Complex problem corrects bias. Result of this unbiasing mechanism in Simple-Complex is responsible for minimizing impact of noisy instances. 
\section{Robust Active Learning}
The objective function (\ref{alfirstobjective}) is convex with respect to $y_s$ and concave with respect to $y_u$. Based on the minimax lemma, they can be exchanged in optimization. Using a binary variable, $q_i \in \{0,1\}$, the problem (\ref{alfirstobjective}) with objective function (\ref{eq:alcostfunction}) replaced with objective (\ref{l01problem}) becomes
\begin{IEEEeqnarray}{lll}
x_s^* =arg\min_{q_s} \max_{y_s} \mathcal{\hat{L}}(\mathcal{D}_l,\mathcal{D}_u,x_s)\\
\mathcal{\hat{L}}(\mathcal{D}_l,\mathcal{D}_u,q_s,y_s)=\min_{p,y_u} \min_{w\in \mathcal{H},w_o\in \mathcal{H}_o}\IEEEnonumber\\ \frac{1}{n}\sum_{i\in D} (1-q_i) (1-y_{i}\langle w_o,\psi(x_i)\rangle)_+ (1-y_{i}\langle w,\phi(x_i)\rangle)_+ \IEEEnonumber \\ 
+\frac{1}{n}\sum_{j\in Q} q_j \big(1-y_{j}\langle w,\psi(x_j)\rangle \big)_+ +\frac{\lambda}{2}\Vert w\Vert_{\mathcal{H}} ^{2} +\frac{\lambda_o}{2}\Vert w_o\Vert_{\mathcal{H}}^{2}\IEEEnonumber\\
\text{s.t.} \forall i \text{  ,  } \vert \langle w_o,\psi(x_i)\rangle \vert \leq p_i \IEEEnonumber \\
 p^\mathsf{T} \mathbf{1} \leq n_o, p_i\in \{0,1\}\IEEEnonumber\\
\sum_{i=1}^n q_i = b ,q_i \in \{0,1\}\IEEEnonumber
\end{IEEEeqnarray}
A noisy instance cannot be selected for querying. Therefore, $p_i+q_i\leq 1$. In this problem, $q_i=0$ for all $x_i\in D-Q$. Using $p_i$ we can compare noisyness of two instance, or add some constraints about noisyness of instances. This can even be used to query degree of noisyness of an instance. 

Unfortunately, this problem is a non-convex integer program and is very difficult to solve. Constraints about domain of variables $p_i$ and $q_i$ are relaxed to $[0,1]$.
\newcommand{\tvect}[2]{
  \ensuremath{\Bigl[\negthinspace\begin{smallmatrix}#1\\#2\end{smallmatrix}\Bigr]}}
\newcommand{\hvect}[3]{%
	\ensuremath{\Bigl[\negthinspace\begin{smallmatrix}#1\\#2\\#3\end{smallmatrix}\Bigr]}}

\begin{theorem}
	\label{relaxpqtheorem}
	If $q$ and $p$ is relaxed in the above problem, then this problem is equivalent to  
	\begin{IEEEeqnarray}{lll}
	\min_{q,y_u,w_o,p} \max_{\alpha} &\sum_{i\in D} \widetilde{\alpha}_i (1-g_i) +\frac{\lambda_o}{2}\Vert w_o\Vert ^{2} -\frac{1}{2\lambda} \alpha^\mathsf{T} \big(K\odot \widehat{G}\big) \alpha \IEEEnonumber\\
	\text{s.t.} &0\leq \widetilde{\alpha} \leq 1, -1\leq \tau \leq 1\\
	& \widehat{G}= \tvect{Y(1-g)}{q} \tvect{Y(1-g)}{q}^\mathsf{T}\label{Gmatrixconstraint}\\
	& \lefteqn{\mathbf{1}-g=(I-diag(q))(\mathbf{1}-Y \Psi(X)^\mathsf{T}w_o)}\\
	& \vert \Psi(X)^\mathsf{T} w_o  \vert \leq p, p_i\in \{0,1\},\mathbf{1}^\mathsf{T} p \leq n_o\\
	&y_{ui}\in \{-1,1\},
	\end{IEEEeqnarray}
\end{theorem}   
This problem is highly non-convex. Before devising a convex-relaxation for this problem, a convex relaxation for the problem without $w_o$ is proposed. In this case the sole source of non-convexity is (\ref{Gmatrixconstraint}). Based on convex-relaxation proposed by Geomans and Williamson \cite{Goemans1995}, this constraint can be relaxed to 
\begin{align}
&\widehat{G}  \succcurlyeq  \tvect{Y(1-q)}{q} \tvect{Y(1-q)}{q}^\mathsf{T},diag(\widehat{G})=\tvect{\mathbf{1}-\hat{q}}{q} 
\end{align}
where $\forall i\in D_l, \hat{q}_i = 0, \forall i\in D_u, \hat{q}_i=q_i$. Based on the result of Geomans and williamson\cite{Goemans1995}, the accuracy of this high(clear the accuracy). Using schur lemma, this is equivalent to 
\begin{IEEEeqnarray}{lcl}
\label{Grelaxed}
G=\left[\begin{IEEEeqnarraybox}{c'c}
\widehat{G}&g\\
g^\mathsf{T}&1
\end{IEEEeqnarraybox}\right] \succcurlyeq 0,\text{where } g=\hvect{y_l}{v_u}{q}\\
diag(\widehat{G})=\mathbf{1}-\hat{q},\forall i\in \mathcal{D}_u, \hat{q}_i = q_i,i\in \mathcal{D}_l, \hat{q}_i = 0  
\end{IEEEeqnarray}
,where $v_u=Y_u(1-q)$. Unfortunately, this constraint is still non-convex. Since $Y_u\in \{-1,1\}$, we have $\vert v_u \vert + q =1$. In this case, convex-relaxation for the problem is
\begin{IEEEeqnarray}{lll}
\min_{q,y_u,G} \max_{\alpha} &\sum_{i\in D} \widetilde{\alpha}_i (1-q_i) -\frac{1}{2\lambda} \alpha^\mathsf{T} \big(K\odot \widehat{G}\big) \alpha \\
\text{s.t.} &0\leq \widetilde{\alpha} \leq 1, -1\leq \tau \leq 1\\
& G \succcurlyeq 0\\
&\vert v_u \vert + q = 1 \\
&diag(\widehat{G})=\mathbf{1}-q,\forall i\in D_l, q_i = 0 \\
&\mathbf{1}^\mathsf{T} q = b, q \in [0,1]
\end{IEEEeqnarray}
where $\widehat{G}$ is as in (\ref{Grelaxed}).
 
For problem in theorem (\ref{relaxpqtheorem}), constraint (\ref{Gmatrixconstraint}) can be written as (\ref{Grelaxed}) but with $g=\hvect{v_l}{v_u}{q}$, where $v_l$, $v_u$ can be written 
\begin{align}
v_l&=Y_l diag(1-Y_l \Psi(X_l)^\mathsf{T}w_o)\\
v_u&= Y_u diag(1-Y_u \Psi(X_u)^\mathsf{T}w_o)(1-q) \cr
\vert v_u \vert &=diag(1-Y_u \Psi(X_u)^\mathsf{T}w_o)(1-q) \cr
&= 1-Y_u \Psi(X_u)^\mathsf{T}w_o-q+diag(q) Y_u \Phi(X_u)^\mathsf{T}w_o 
\end{align} 
Since we don't want to select noisy instances for querying, at least one of $q_i$ or $\Psi(X_u)^\mathsf{T}w_o$ is very small and the other one is less than one, therefore the last term is small.  So, the last equation is just becomes $\vert v_u \vert+ q + Y_u \Psi(X_u)^\mathsf{T} w_o =1$. Assuming $p_i = Y_u \Psi(X_u)^\mathsf{T}w_o$ is a good approximation, since $w_o$ is complex therefore it can more easily fit data.  Using this approximation, this equation becomes
\begin{align}{\label{eq:YuPhiw_o}}
\vert v_u \vert+ q + Y_u \Psi(X_u)^\mathsf{T} w_o =1\\
\vert \Psi(X_u)^\mathsf{T}w_o \vert +q <=1\notag
\end{align}
   
Instead using the following approximation may produce better results 
\begin{IEEEeqnarray}{lcl}
r=Y_u\Psi(X_u)^\mathsf{T} w_o\\
v_u = Y_u diag(1-Y_u \Psi(X_u)^\mathsf{T}w_o)(1-q) \\
\Psi(X)^\mathsf{T}w_o = p^+-p^-\\
\vert v_u \vert+ q + r =1\\
\Vert r \Vert_1 \leq n_o \textbf{, or} \Vert r \Vert_1 \leq  \sum_i(p^+_i+p^-_i)\\
v_u+\Psi(X_u)^\mathsf{T}w_o=Y_u ( 1-q) \equiv r_q\\
-(1+Y_u)<= r+\Psi(X_u)^\mathsf{T} w_o <= 1+Y_u\\
-1+Y_u  <= r-\Psi(X_u)^\mathsf{T} w_o <= 1-Y_u\\
-1+q <= r_q <= 1-q\\
-q+Y_u <= r_q <= q+Y_u\\
-(1+Y_u)<= r_q+1-q <= 1+Y_u\\
-1+Y_u  <= r_q-1+q <= 1-Y_u
\end{IEEEeqnarray}
In addition to the above constraints, we must add, $c\sum_i(p^+_i+p^-_i)$ to objective function to enforce $p=\vert \Psi(X)^\mathsf{T} w_o\vert$. 
Furthermore we have 
  \begin{IEEEeqnarray}{lll}
  \mathbf{1}-g&=(I-diag(q))(\mathbf{1}-Y \Psi(X)^\mathsf{T}w_o)\\
  &=\mathbf{1} - q -Y \Psi(X)^\mathsf{T}w_o +diag(q)) Y \Psi(X)^\mathsf{T}w_o 
  \end{IEEEeqnarray}
  Therefore, 
  \begin{IEEEeqnarray}{lcl}
  g&= q +Y \Psi(X)^\mathsf{T}w_o -diag(q)  Y \Psi(X)^\mathsf{T}w_o\\
  &= 1-\vert v \vert
  \end{IEEEeqnarray}
The final form for this problem using \ref{eq:YuPhiw_o} is 
\begin{IEEEeqnarray}{lll}
\label{eq:FinalRAL}
\min_{G,q,y_u,w_o,p}\max_{\alpha} 
&\sum_{i\in D} \widetilde{\alpha}_i (1-g_i)  
-\frac{1}{2\lambda} \alpha^\mathsf{T} (K\odot \hat{G})\alpha +c_a \mathbf{1}^\mathsf{T} a\IEEEnonumber\\
&+\frac{\lambda_o}{2}\Vert w_o\Vert ^2\\
\textrm{s.t.} &0\leq \widetilde{\alpha} \leq 1, -1\leq \tau \leq 1\\
&diag(\widehat{G}) = \tvect{\mathbf{1}-g}{q} \\
&g = 1-\vert v\vert \\
&\vert v \vert \leq a\\
&a+ q + p =1\\
&\vert \Psi(X_u)^\mathsf{T}w_o \vert <=1-q\\
& \vert \Psi(X)^\mathsf{T} w_o  \vert \leq p\\
& \mathbf{1}^\mathsf{T} p \leq n_o\\
& \mathbf{e_l}^\mathsf{T} p <= n_{lbn}\\
& \mathbf{1}^\mathsf{T} q = b\\
& q,p\in [0,1]
\end{IEEEeqnarray}
Based on representation lemma, $w_o=\Phi(X)\beta $. For notational simplicty, in objective function (\ref{eq:FinalRAL}) , let $x=(u,\beta,s)$ and $u=(G,p)$ .(If linear approximation for $Y_u\Psi(X_u)^\mathsf{T}w_o$ is used define $u=(G,p,a,g,r,y_u)$ ). Then constraint set of the above problem can be represented using proper definition of the operators $A_{EC}$, $A_{EV}$, $B_{EV}$, $A_{IC}$, $A_{IV}$,$B_{IV}$ as 
\begin{IEEEeqnarray}{lll}
	\label{eq:constraintset}
\mathcal{C} =\Big\{x=(u,\beta,s) \textbf{\big|} &A_{EC}(u)= b_{EC},A_{EV}(u)= b_{EV} + B_{EV} \beta\IEEEnonumber\\
&,A_{IC}(u) = s_{IC}, A_{IV}(u) = s_{IV} + B_{IV} \beta\IEEEnonumber\\
&,u \in \mathcal{S}_+^n \times \mathcal{R}_+^n\times  \mathcal{R}_+^n \times \mathcal{R}_+^n,u \in \mathcal{P},\IEEEnonumber\\
& s=\tvect{s_{IC}}{s_{IV}} \in \mathcal{K}\Big\}
\end{IEEEeqnarray} 
In this way, the problem becomes more of a standard conic problem:
\begin{align}\label{eq:StandardFinalForm}
&\min_{x\in \mathcal{C}}  \max_{\alpha} -f(x,\alpha) 
\end{align}
where 
\begin{align}\label{eq:Functionform}
&f(x,\alpha)=-a^\mathsf{T} \alpha -\frac{\lambda_o}{2} \beta^\mathsf{T} K \beta 
+\frac{1}{2\lambda} \alpha^\mathsf{T} (K\odot \hat{G})\alpha -c_a \mathbf{1}^\mathsf{T} a 
\end{align}
and we have 
\begin{align}
&\bigtriangledown_{\alpha} f(x,\alpha)= -a 
+\frac{1}{\lambda}  (K\odot \hat{G})\alpha 
\end{align}
with some simplification about $a$ and $g$ 
\begin{align}
a &= 1- p -q \cr g &= 1-a= 1-(1-p-q)=p+q
\end{align}
we have
\begin{IEEEeqnarray}{lll}
\label{eq:FinalRAL2}
&-f(x,\alpha)\\
&=\sum_{i\in D} \widetilde{\alpha}_i (1-p_i-q_i) +\frac{\lambda_o}{2}\Vert w_o\Vert ^2 
-\frac{1}{2\lambda} \alpha^\mathsf{T} (K\odot \hat{G})\alpha\IEEEnonumber\\
&+c_a \mathbf{1}^\mathbf{T} \mathbf{1} - c_a \mathbf{1}^\mathsf{T} (p+q)\\
&=\mathbf{1}_D^\mathsf{T} \alpha -\widetilde{\alpha}^\mathsf{T}(p+q) +\frac{\lambda_o}{2}\Vert w_o\Vert ^2 
-\frac{1}{2\lambda} \alpha^\mathsf{T} (K\odot \hat{G})\alpha\IEEEnonumber\\
&+c_a \mathbf{1}^\mathbf{T} \mathbf{1} - c_a \mathbf{1}^\mathsf{T} (p+q)\\
&=\mathbf{1}_D^\mathsf{T} \alpha -(\widetilde{\alpha}+c_a \mathbf{1})^\mathsf{T}(p+q) +\frac{\lambda_o}{2}\Vert w_o\Vert ^2 
-\frac{1}{2\lambda} \alpha^\mathsf{T} (K\odot \hat{G})\alpha\IEEEnonumber\\
&+c_a \mathbf{1}^\mathbf{T} \mathbf{1}
\end{IEEEeqnarray}
And finally, 
\begin{IEEEeqnarray}{lll}
-f(x,\alpha)= &\mathbf{1}_D^\mathsf{T} \alpha -(\widetilde{\alpha}+c_a \mathbf{1})^\mathsf{T}(p+q) +\frac{\lambda_o}{2} \beta^\mathsf{T} K \beta \IEEEnonumber\\
&-\frac{1}{2\lambda} \alpha^\mathsf{T} (K\odot \hat{G})\alpha 
\end{IEEEeqnarray}
\subsection{Solving Robust Active Learning Problem}
Function $f(x,\alpha)$ is concave with respect to $x$ and convex with respect to $\alpha$ and constraint set of the problem (\ref{eq:StandardFinalForm}), $\mathcal{C}$ are affine. Based on the minimax lemma, $\max_x \min_{\alpha} f(x,\alpha)=\min_{\alpha} \max_x f(x,\alpha)$. Therefore this problem is a convex-concave saddle point problem. It is well known that the operator defined as $T(x,y)=\partial \big(f(.,\alpha)-f(x,.)\big)(x,\alpha)$ for this problem is maximal monotone. If $(0,0)\in T(x,\alpha)$, then $(x,\alpha)$ is the saddle point of the problem (\ref{eq:Functionform}). We proposed two method for this problem. The first method is based on forward-backward-forward method or Tseng's method\cite{Bauschke}.
\subsubsection{forward-backward-forward method}  
By building a maximal monotone operator $F$ and finding its fixed point, i.e.,  $x^*=F(x^*)$ for problem (\ref{eq:StandardFinalForm}), saddle point of the problem can be obtained. By corollary 24.5 in \cite{Bauschke}, the following operator $T(x,\alpha)$ is maximally monotone.
\begin{align}
T(x,\alpha)&=\partial \big(f(.,\alpha)-f(x,.)\big)(x,\alpha)\cr
&=A+B\cr
A&=\partial \big(f(.,\alpha))(x,\alpha)\cr
B&=\partial \big(-f(x,.)\big)(x,\alpha)
\end{align}
Based on theorem 25.10 in \cite{Bauschke} using Tseng's method, and defining $ t_{x}^n=(x,\alpha)^n$ the following iteration converges to saddle point of the above problem
\begin{align}
t_y^n &= t_x^n - \gamma Bt_x^n\cr
t_z^n &= J_{\gamma A} t_y^n\cr
t_r^n &= t_z^n - \gamma Bt_z^n\cr
t_x^{n+1} &= P_C(t_x^n-t_y^n+t_r^n)
\end{align}	
For $t_y^n = t_x^n - \gamma Bt_x^n$ we have:
\begin{IEEEeqnarray}{lll}
-B(\alpha) &=-\partial \big(-f(x,.)\big)(\alpha) = (1_D-g_D)- \frac{1}{\lambda} (K\odot \widehat{G}) (\alpha)\IEEEnonumber\\
\alpha_y^n &= \alpha_x^n - \gamma Bt_x^n\IEEEnonumber
\end{IEEEeqnarray}
and for $ t_z^n = J_{\gamma A} t_y^n $:  
\begin{IEEEeqnarray}{lll}
\label{SDPsubproblem}
x_z^n &= arg\min_{x \in \mathcal{C}} \big\{ -\langle c_k,u \rangle +\frac{\lambda_o}{2} \beta^\mathsf{T} K \beta+\rho/2 \Vert x-x^k\Vert_{\mathcal{S}}^2 \big\}\IEEEnonumber\\
t_z^n &= (x_z^n,\alpha^{n}),\rho=1/\gamma
\end{IEEEeqnarray}
where 
$c_k=(\frac{1}{2\lambda} K\odot \alpha^k \alpha^{k\mathsf{T}}+\sum_{i\in q}A_{q_i}(\alpha_i+c_a), \widetilde{\alpha}+c_a \mathbf{1})$ and $\langle , \rangle$ is the proper inner product for $u$ and $A_{q_i}$ is an operator such that $A_{q_i}(u)=q_i$. We define $\mathcal{S}$ as  
\begin{align}
&\Vert x-x^k\Vert^2=\Vert u-u^k\Vert^2+\Vert s-s^k \Vert^2 + (\beta-\beta^k)^\mathsf{T} Q (\beta-\beta^k)
\end{align}
This is proximal point step in a conic space. Proof of the following theorem which is based on \cite{Sun2016} can be found in Supplementary   Material.
\begin{theorem}
	\label{innersdpproblem}
	Dual of the above problem is 
	\begin{IEEEeqnarray}{lll}
	\min_{S,v,Z,y} &\frac{1}{2\rho} \Vert  A^*( y)+c_k+S+Z +\rho u^k\Vert^2-\frac{\rho}{2} \Vert u^k\Vert^2\IEEEnonumber\\
	&-\langle  y, b_E\rangle +\frac{1}{2} \Vert B^*(y) +\rho \beta^{k\mathsf{T}} Q \Vert^2_{R} -\frac{\lambda_o}{2} \beta^{k\mathsf{T}} K \beta^k\IEEEnonumber\\
	&+\frac{1}{2\rho} \Vert  v+y_I-\rho s^k \Vert^2 -\frac{\rho}{2}\Vert s^k\Vert^2+\langle v, s_I\rangle  \IEEEnonumber\\
	&S\in S_+^n, v\in \mathcal{K}^*, Z \in \mathcal{P}^*
	\end{IEEEeqnarray}
\end{theorem}
where $R=( \lambda_o K + \rho Q)^{-1}$.
Derivation with respect to $y_E$ and $y_I$ is 
\begin{align}
\frac{\partial \mathcal{L}}{\partial y_E} = &\frac{1}{\rho} A_E (\rho u^k+ A_E^*( y_E)+A_I^*( y_I)+c_k+S+Z)\cr&-b_E+B_E R (B_E^*y_E+B_I^*y_I+\rho \beta^{kT}Q)\nonumber\\
\frac{\partial \mathcal{L}}{\partial y_I} = &\frac{1}{\rho} A_I (\rho u^k+ A_E^*( y_E)+A_I^*( y_I)+c_k+S+Z)\cr&+B_I R (B_E^*y_E+B_I^*y_I+\rho \beta^{kT}Q)+\frac{1}{\rho}(v+y_I-\frac{s^k}{\rho})\nonumber
\end{align}
and setting zero, we will have 
\begin{align}
(A_E A_E^*+\rho B_E R B_E^*) y_E &= \rho b_E- A_E (A_I^*y_I+c_{SZu})\cr
&-\rho B_E R (B_I^*y_I+\rho \beta^{kT}Q)\cr
( A_I A_I^*+\rho B_I R B_I^*+I) y_I= &\frac{s^k}{\rho}-v - A_I (A_E^*y_E+c_{SZu})\cr&-\rho B_I R (B_E^*y_E+\rho \beta^{kT}Q)
\end{align}
where $c_{SZu}=c_k+S+Z+\rho u^k$. 
KKT condition for primal problem 
\begin{align}
A_{EC}(u)&= b_{EC}\cr
A_{EV}(u)&= b_{EV} - B_{EV} \beta\cr
A_{IC}(u)&= s_{IC}\cr
A_{IV}(u)&= s_{IV} - B_{IV} \beta \cr
u - u^{\prime} &= 0\cr
u &= \Pi_{(S^n_+,R^{n+})}(u^k+\frac{1}{\rho}(A^*( y)+c_k+Z))\cr
u^{\prime} &= \Pi_{\mathcal{P}}(u^k+\frac{1}{\rho}(A^*( y)+c_k+S))\cr
s &= \Pi_{\mathcal{K}}(s^k-\frac{y_I}{\rho}) 
\end{align}
\subsection{Using Nesterov's Method for Composite Functions}
From previous iteration of the (active) learning algorithm, we a have an starting point for the optimization problem. Assume $x_{st}$ and $\alpha_{st}$ are these starting point. Considering the following regularized convex-concave problem: 
\begin{align}
min_{\alpha \in Q} max_{x\in P} f(x,\alpha)+ \frac{\rho}{2} \Vert\alpha-\alpha_{st}\Vert^2-\frac{\rho}{2} \Vert x-x_{st} \Vert^2	
\end{align}
Define 
\begin{align}
f_{\rho} (\alpha)=\max_{x\in P} \Big\{f(x,\alpha)-\frac{\rho}{2} \Vert x-x_{st} \Vert^2\big\},
\Psi_{\rho}(\alpha) = \frac{\rho}{2} \Vert\alpha-\alpha_{st}\Vert^2
\end{align}
Function $f_{\rho}(\alpha)$ has derivative and $\Psi_{\rho}(\alpha)$ is a closed-convex function. Therefore we can use Nesterov's method for composite functions \cite{Nesterov2013}. This is suggested in many researches including \cite{Devolder2014}, \cite{Kolossoski2015}. Based on  Danskin theorem \cite{Sun2016} we have 
\begin{align}
\bigtriangledown_{\alpha} f_{\rho}(\alpha^\prime) = \bigtriangledown_{\alpha} f(x,\alpha^\prime) =  a(\alpha^\prime)- \frac{1}{\lambda} (K\odot \widehat{G}(\alpha^\prime)) (\alpha^\prime)
\end{align}
If there exists constants $L{\alpha \alpha}, L_{\alpha x} $ such that 
\begin{align}
\Vert \bigtriangledown_{\alpha}f(x^\prime,\alpha^\prime) -\bigtriangledown_{\alpha}f(x,\alpha) \Vert \leq L_{\alpha \alpha} \Vert \alpha^\prime-\alpha \Vert + L_{\alpha x } \Vert x-x^\prime \Vert
\end{align}
Since function $f(.,\alpha)$ is a $\rho$-stronly convex function based on \cite{Monteiro2010} we have 
\begin{align}
f_{\rho}(\alpha) \leq f_{\rho}(\alpha^\prime )+ \langle g, \alpha-\alpha^\prime \rangle + \frac{L}{2} \Vert \alpha-\alpha^\prime \Vert^2+ \delta 
\end{align}
where $L=2\big( L_{\alpha \alpha}+\frac{L_{\alpha x}^2}{\rho}\big) ,g \in \partial_{\delta} f_{\rho}(\alpha^\prime)$. (Obtaining Lipschitz constant for this problem and using danskin's theorem we can convert all of this to a problem). 

Now based on the method Nesterov's method \cite{Nesterov2013} consider function $\phi_{\rho}(\alpha) = f_{\rho}(\alpha) + \Psi_{\rho}(\alpha)$. Define 
\begin{IEEEeqnarray}{lll}
m_L(\alpha_0;\alpha) = &f_{\rho}(\alpha_0) + \langle \bigtriangledown f_{\rho}(\alpha), \alpha-\alpha_0 \rangle\IEEEnonumber\\
&+ \frac{L}{2} \Vert \alpha-\alpha_0 \Vert^2 + \Psi_{\rho}(\alpha),\\
T_L(\alpha_0)=&arg\min_{\alpha\in P} m_L(\alpha_0;\alpha)  
\end{IEEEeqnarray}
and 
\begin{IEEEeqnarray}{lll}
	\label{psikcomp}
\psi_0(\alpha)&=\frac{1}{2} \Vert \alpha-\alpha_0\Vert^2,\IEEEnonumber\\
\psi_{k}(\alpha)&= \psi_{k-1}(\alpha) + a_{k} [f_{\rho}(\alpha_{k})+\langle \bigtriangledown_{\alpha} f_{\rho}(\alpha_{k}),\alpha-\alpha_{k}\rangle + \Psi_{\rho}(\alpha)]\IEEEnonumber\\
&=\frac{1}{2} \Vert \alpha- \alpha_{0} \Vert^2 + \sum_{i\in [1,k]} a_{i} f_{\rho}(\alpha_{k}) \IEEEnonumber\\
&+ \sum_{i\in [1,k]} a_{i} \bigtriangledown_{\alpha}f_{\rho}(\alpha_{i}) \alpha + \sum_{i\in [1,k]} a_{i} \Psi_{\rho}(\alpha)
\end{IEEEeqnarray}
Considering $d(\alpha, \alpha_{0}) = \Vert \alpha-\alpha_{0}\Vert^2$
\begin{align}
\psi_{k}(\alpha)&=\frac{1+A_k}{2} \Vert \alpha- \alpha_{0} \Vert^2 + \sum_{i\in [1,k]} a_{i} \bigtriangledown_{\alpha}f_{\rho}(\alpha_{i}) \alpha 
\end{align}
Therefore $\bigtriangledown_{\alpha}\psi_k(\alpha) = (1+A_k)(\alpha-\alpha_{0}) + \frac{2}{1+A_k} \sum_{i\in [1,k]} a_{i} \bigtriangledown_{\alpha}f_{\rho}(\alpha_{i})$.
The final algorithm for problem is Algorithm (~\ref{alg:SaddleNestrovAccelerated}).
\begin{algorithm}
	\caption{Nesterov's Accelerated Composition Method} 
	\label{alg:SaddleNestrovAccelerated}
	\begin{algorithmic}[1]
		\State Input : $L_{\alpha \alpha}, L_{\alpha x},\alpha_{0},x_{0}$
		\State Output : $\beta_K,\hat{x}_K$
		\State $\psi_0(\alpha) = \frac{1}{2} \Vert \alpha- \alpha_{st} \Vert^2,A_0 =0,v_0 =\alpha_0,\beta_{0}=\alpha_0$
		\For{\texttt{k=1,2,...K}}
		\State $a \gets \text{positive root of } \frac{a^2}{A_k+a}=\frac{1+\mu A_k}{L/2}$
		\State $\alpha_k \gets \frac{A_k}{A_k+a} \beta_{k-1} + \frac{a}{A_k+a} v_{k-1}$
		\State $x_k= argmin\{-f(x,\alpha_k)+\frac{\rho}{2}\Vert x-x_{0}\Vert_H^2\}$
		\State Compute $f(x_k,\alpha_{k}),\bigtriangledown_{\alpha} f(x_k,\alpha_{k})$
		\State Update $\psi_{k}(\alpha)$ using (\ref{psikcomp})
		\State $ v_{k} \gets argmin \{  \psi_{k}(\alpha)   \}$
		\State $\hat{x}_k \gets \frac{A_k}{A_k+a}\hat{x}_{k-1}+ \frac{a}{A_k+a} x_k$
		\State $\beta_{k}\gets \frac{A_k}{A_k+a} \beta_{k-1}+\frac{a}{A_k+a} v_k$
		\State $A_{k+1} \gets A_k+a$
		\EndFor
	\end{algorithmic}
\end{algorithm}
\section{Conclusion}
In this paper, a new active learning method is proposed. It is based on the instance complexity. Instance complexity is a measure for noisiness of instances with respect to learning. It is shown that usefull instances are the ones in a range of "complexy"-ties. This is very intuitive as it relates the concept of usefullness of instances in a learning problem to the concept of complexity which has a central role in understanding machine learning algorithms. The proposed loss function and the resulting Simple-Complex classifier uses only usefull data points for learning. It is shown that the resulting algorithm is unbiased. Furthermore, it is very important that accuracy of representativeness and informativeness as the two most common active learning methods is inherently related to noisy instances in the data. Two methods for solving the problem is proposed. In future, we plan to work on improving speed of the algorithm and extending this work to distributional data.


%

\ifCLASSOPTIONcompsoc
  \section*{Acknowledgments}
    
\else
  \section*{Acknowledgment}
\fi
\ifCLASSOPTIONcaptionsoff
  \newpage
\fi


%
\bibliographystyle{IEEEtran}
\bibliography{RobustIEEEALPaper}
%

%

\begin{IEEEbiography}{Hossein Ghafarian}
received the B.S. degree in computer engineering from Isfahan University of Technology, Isfahan, Iran, in 1996. And M.S. degree in computer engineering from Ferdowsi University of Mashhad, Iran, in 1999. He is a faculty member of Quchan Institute of Engineering and Technology. Also, He is currently pursuing a Ph.D. degree in computer engineering with Ferdowsi University of Mashhad, Iran, department of Computer Engieerning, where he is a member of Pattern Recognition and Machine Learning Laboratory. His main research interest includes active learning and its various applications in Machine Learning, convex methods for combinatorial optimization, learning from distributional data and causality. 
\end{IEEEbiography}

\begin{IEEEbiography}{Hadi Sadoghi Yazdi}
was born in Sabzevar, Iran, in 1971. He received his B.S. in Electrical Engineering
from Ferdowsi University of Mashhad (FUM) 1994, and received his M.S. and Ph.D in Electrical
Engineering from Tarbiat Modares University Tehran in 1996 and 2005, respectively. Currently he is a Professor at FUM. Dr. Sadoghi Yazdi has received several awards including Outstanding
Faculty Award from Iran Ministry of Science, Research and Technology (MSRT) in 2009, 2010, 2011
and Best System Design Award from Ferdowsi Festival in 2007. His research interests are pattern
recognition, machine learning, machine vision, signal processing, data mining and optimization
\end{IEEEbiography}





\end{document}